\documentclass[sigconf]{acmart}

\AtBeginDocument{%
  \providecommand\BibTeX{{%
    \normalfont B\kern-0.5em{\scshape i\kern-0.25em b}\kern-0.8em\TeX}}}

\setcopyright{rightsretained}
\copyrightyear{2022}
\acmYear{2022}
\acmConference{SIGGRAPH '22 Posters}{August 07-11, 2022}{Vancouver, BC, Canada}\acmBooktitle{Special Interest Group on Computer Graphics and Interactive Techniques Conference Posters (SIGGRAPH '22 Posters), August 07-11, 2022}\acmDOI{10.1145/3532719.3543244}
\acmISBN{978-1-4503-9361-4/22/08}

\begin{document}

\title{Generating Diverse Indoor Furniture Arrangements}

\author{Ya-Chuan Hsu}
\email{yachuanh@usc.edu}
\affiliation{%
  \institution{University of Southern California}
  \city{Los Angeles}
  \country{USA}
}

\author{Matthew C. Fontaine}
\email{mfontain@usc.edu}
\affiliation{%
  \institution{University of Southern California}
  \city{Los Angeles}
  \country{USA}
}

\author{Sam Earle}
\email{se2161@nyu.edu}
\affiliation{%
  \institution{New York University}
  \city{New York}
  \country{USA}
}

\author{Maria Edwards}
\email{mae236@nyu.edu}
\affiliation{%
  \institution{New York University}
  \city{New York}
  \country{USA}
}

\author{Julian Togelius}
\email{julian@togelius.com}
\affiliation{%
  \institution{New York University}
  \city{New York}
  \country{USA}
}

\author{Stefanos Nikolaidis}
\email{nikolaid@usc.edu}
\affiliation{%
  \institution{University of Southern California}
  \city{Los Angeles}
  \country{USA}
}

\renewcommand{\shortauthors}{Hsu, et al.}

\begin{abstract}

We present a method for generating arrangements of indoor furniture from human-designed furniture layout data. Our method creates arrangements that target specified diversity, such as the total price of all furniture in the room and the number of pieces placed. To generate realistic furniture arrangement, we train a generative adversarial network (GAN) on human-designed layouts. To target specific diversity in the arrangements, we optimize the latent space of the GAN via a quality diversity algorithm to generate a diverse arrangement collection. Experiments show our approach discovers a set of arrangements that are similar to human-designed layouts but varies in price and number of furniture pieces.

\end{abstract}


\maketitle
\section{Introduction}

Scene synthesis has been heavily studied by the graphics community with applications in AR/VR gaming and autonomous agent or vehicle testing. Our work focuses on the problem of arranging indoor furniture, which has applications in testing robots deployed in homes. Existing methods rely on learning relative positions of furniture from a database~\cite{wang2018deep}, grouping objects by common human activity~\cite{fu2017adaptive}, or completing partially arranged floor plans~\cite{paschalidou2021atiss}.

We propose to generate collections of arrangements that vary according to desired attributes specified by a designer, rather than generating solutions with diverse spacial relations. For example, an interior designer may propose a collection of layouts that are affordable for a variety of budgets. Our method generates a collection of furniture layouts via a three stage process. First, we train StyleGAN3~\cite{karras2021alias} on furniture occupancy grids to match furniture layout patterns of human designers. To recover 3D objects from the generated occupancy grid, we run a geometric sweep algorithm that fits bounding rectangles to labelled pixels. Finally, we search the latent space of StyleGAN3 with a quality diversity algorithm to produce a collection of arrangements that vary according to price and number of objects in the scene.

\section{Encoding furniture arrangement floor plans}

To generate diverse floor plans, we need to train a generative model that is capable of producing novel arrangements, stylistically matching the layouts from human designers. As a generative model, we choose StyleGAN3, a GAN capable of generating photorealistic images. To generate arrangements instead of images, we train StyleGAN3 to produce occupancy grids that model floor locations containing specific objects. To treat occupancy grids as images in the StyleGAN3 training pipeline, we color-code the furniture categories, such as tables, chairs, storage, and beds, of each object. However, assigning arbitrary colors makes learning occupancy grids more challenging for StyleGAN3, since conceptually similar categories may be far apart in color space. Hence, we learn a 3D embedding of furniture categories that we map to RGB color values.

To produce a 3D color embedding of each category, we first map each category name to a 512-dimensional vector embedding via the Universal Sentence Encoder~\cite{cer2018universal}. We then reduce the encoding to a 3D embedding using t-SNE~\cite{van2008visualizing}. Finally, we scale each channel to the range $[0, 255]$ to match an RGB color. We assign colors to 25 umbrella categories representing a total of over a thousand categories of furniture.

\section{Recovering 3D furniture locations and orientations}

StyleGAN3 is capable of generating novel occupancy grids that match the layouts designed by humans. However, the generated occupancy grids produced are imperfect. To recover 3D furniture locations and orientations from the grid, we run a rectangle sweep algorithm to find the best bounding rectangle for a given color and subregion. The algorithm minimizes the cost (Euclidean distance in color space) of changing all pixel colors within the rectangle to the RGB color code of the furniture category (Fig.~\ref{fig:result_archive}).

\section{Generating diverse arrangements}

While StyleGAN3 can generate arrangements stylistically similar to our training data, we wish to generate arrangements at varying price points and with different quantities of furniture. We can frame this problem as a latent space illumination (LSI) problem~\cite{fontaine2020illuminating}, where we search the latent space of StyleGAN3 via quality diversity (QD) optimization~\cite{pugh2015confronting}. To search latent space via QD, we must specify both objective and measure functions. For the objective, we minimize the repair cost of transforming occupancy grids into furniture locations. This objective ensures final arrangements match the occupancy grid as closely as possible. For measure functions, we compute the price of all furniture extracted from the arrangement and the number of furniture included in the final arrangement. We solve the LSI problem with the state-of-the-art QD algorithm, Covariance Matrix Adaptation MAP-Elites (CMA-ME)~\cite{fontaine2020covariance}.

\section{Experimental results}
\begin{figure}[h]
    \begin{minipage}{\linewidth}
    \centering
    \includegraphics[width=.9\linewidth]{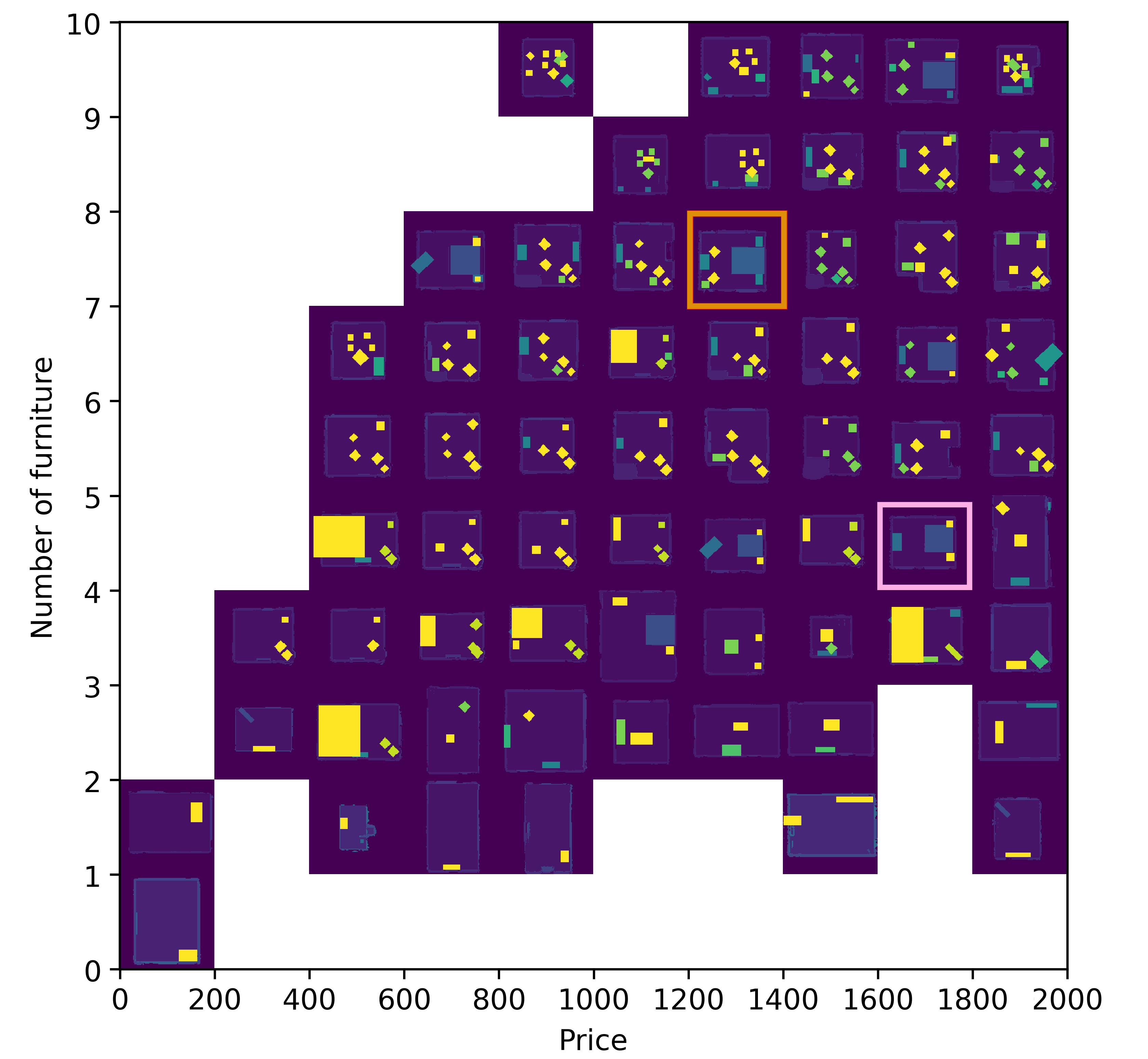}
    \end{minipage}
    \begin{minipage}{.33\linewidth}
    \centering
    \includegraphics[width=.9\linewidth]{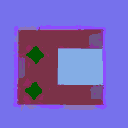}
    \end{minipage}%
    \begin{minipage}{.33\linewidth}
    \centering
    \includegraphics[width=.9\linewidth]{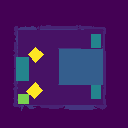}
    \end{minipage}
    \begin{minipage}{.33\linewidth}
    \centering
    \includegraphics[width=.9\linewidth]{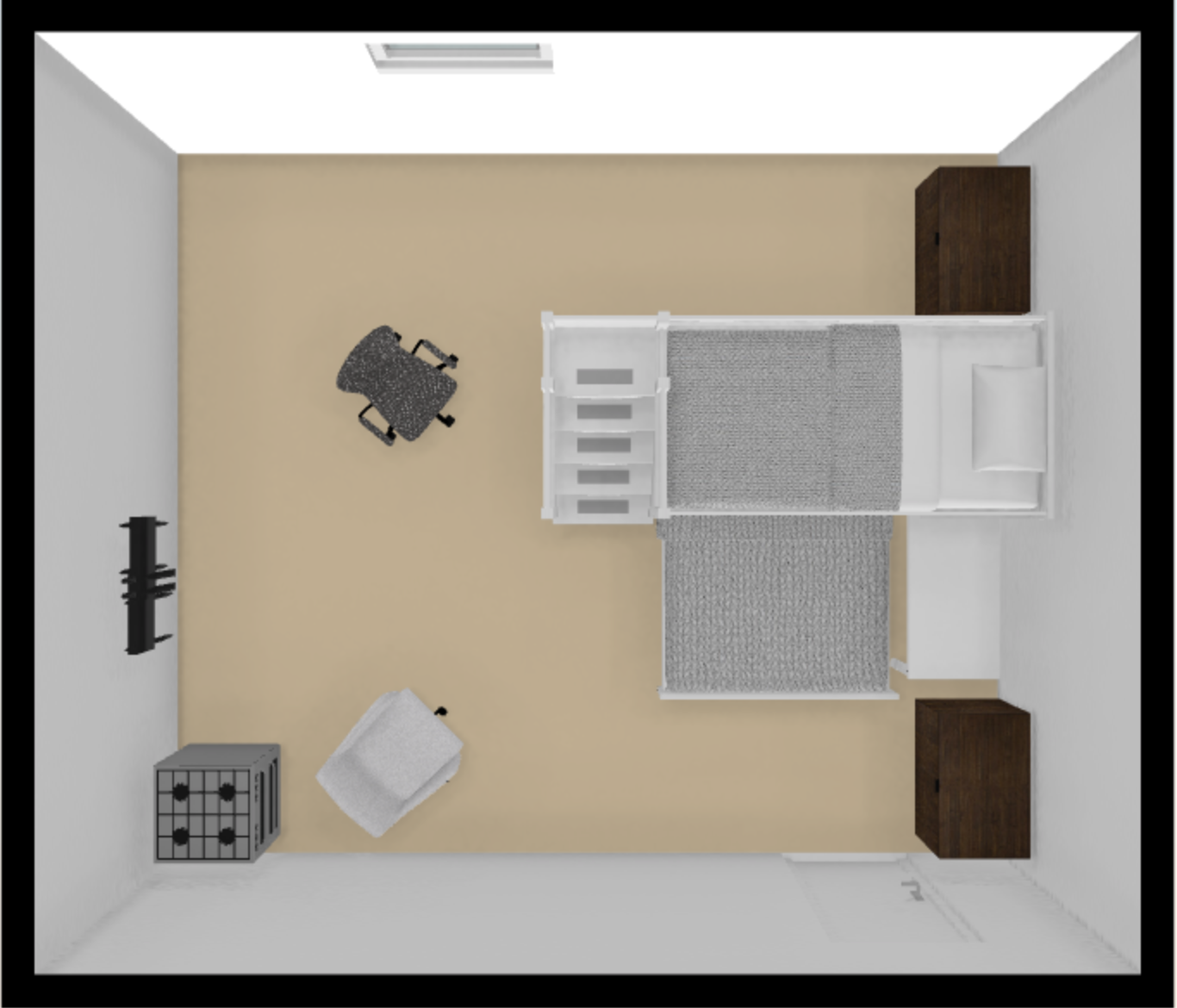}
    \end{minipage}
    \caption{Top: Arrangements in the archive. Bottom: A generated arrangement (left) stored in the orange cell is repaired (middle) and visualized in 3D (right).}
    \label{fig:result_archive}
\end{figure}

We verify our approach by generating a collection of diverse arrangements. Fig.~\ref{fig:result_archive} visualizes the archive of occupancy grids generated by StyleGAN3 and an extracted 3D scene from the occupancy grid. The top-right corner of the archive contains arrangements with a high number of expensive furniture pieces. The bottom-left corner is an empty room. Note that there are empty cells in the archive since it is impossible to have empty rooms with a high price or a room with several furniture pieces with a low price.

\begin{figure}[h]
    \begin{minipage}{.33\linewidth}
    \centering
    \includegraphics[width=.9\linewidth]{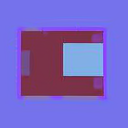}
    \end{minipage}
    \begin{minipage}{.33\linewidth}
    \centering
    \includegraphics[width=.9\linewidth]{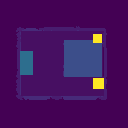}
    \end{minipage}
    \begin{minipage}{.3\linewidth}
    \centering
    \includegraphics[width=.95\linewidth]{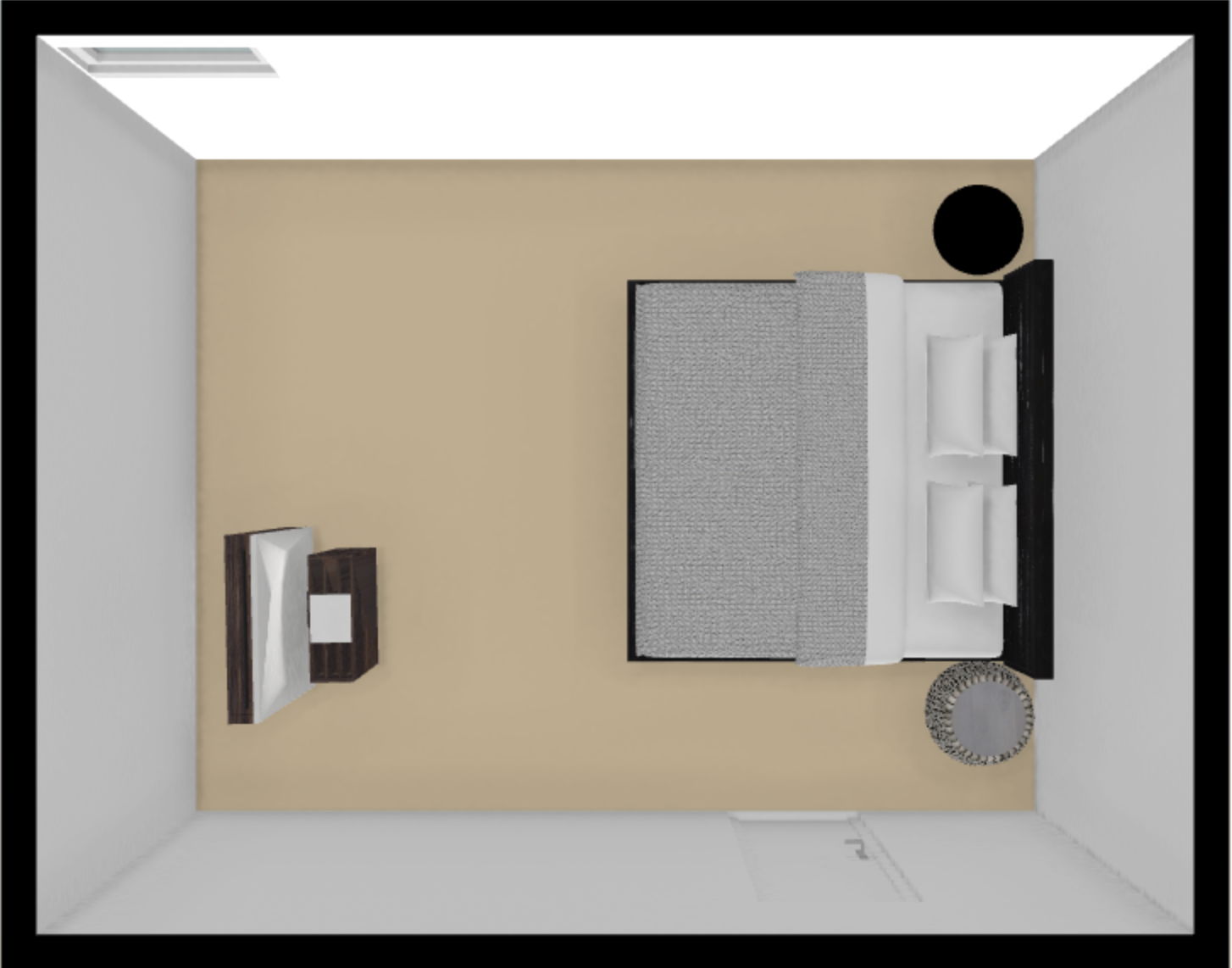}
    \end{minipage}
    \caption{Arrangement in the pink cell from Fig.~\ref{fig:result_archive}}
    \label{fig:comp_price}
\end{figure}

We compare the arrangements in the orange and pink cell (see Fig.~\ref{fig:comp_price}). Even though the orange cell has more furniture pieces, the pink cell arrangement is more expensive due to the `designer walnut vanity unit' and the `California king wood bed'.

\section{Limitations and future work}

An important challenge for our method is extracting 3D furniture locations and orientations from the occupancy grid generated by StyleGAN3. Due to imperfect occupancy grids, extracting furniture locations may result in overlapping furniture. Future repair methods will also aim to address issues such as the stove created in Fig.~\ref{fig:result_archive}, a result of a large block of noise. Additionally, we intend to extend arrangement features to include room accessibility by guaranteeing all furniture is accessible~\cite{fontaine2021importance,zhang2020video}.

\begin{acks}
We thank Floorplanner for providing indoor furniture data. We thank Gert-Jan van der Wel, Valentin Bichkovsky, Vincent van Gemert for assisting with the data and the Floorplanner API.
\end{acks}

\bibliographystyle{ACM-Reference-Format}
\bibliography{main}

\end{document}